\pdfoutput=1
\documentclass[11pt]{article}
\usepackage[ruled,vlined]{algorithm2e}
\usepackage{acl}
\usepackage{times}
\usepackage{latexsym}

\usepackage[T1]{fontenc}
\usepackage[utf8]{inputenc}
\usepackage{amssymb}

\usepackage{microtype}

\usepackage{inconsolata}

\usepackage{graphicx}%
\usepackage{multirow}%
\usepackage{amsmath,amssymb,amsfonts}%
\usepackage{amsthm}%
\usepackage{mathrsfs}%
\usepackage[title]{appendix}%
\usepackage{xcolor}%
\usepackage{colortbl}%
\usepackage{textcomp}%
\usepackage{manyfoot}%
\usepackage{subcaption}
\usepackage{booktabs}%
\usepackage{pifont}
\usepackage{tabularx}
\usepackage{tcolorbox}
\tcbuselibrary{skins}
\usepackage{listings}%

\usepackage{tikz}
\usetikzlibrary{shapes.geometric, arrows.meta}

\definecolor{LavenderBlush}{rgb}{1.0, 0.94, 0.96}
\definecolor{Lavender}{rgb}{0.9, 0.9, 0.98}
\definecolor{MistyRose}{rgb}{1.0, 0.89, 0.88}
\definecolor{MintCream}{rgb}{0.96, 1.0, 0.98}
\definecolor{AliceBlue}{rgb}{0.94, 0.97, 1.0}
\definecolor{Seashell}{rgb}{1.0, 0.96, 0.93}
\definecolor{LightYellow}{rgb}{1.0, 1.0, 0.88}
\definecolor{Peach}{rgb}{1.0, 0.9, 0.71}
\definecolor{Apricot}{rgb}{0.98, 0.81, 0.69}
\definecolor{LightCyan}{rgb}{0.88, 1.0, 1.0}
\definecolor{rose}{rgb}{1.0, 0.0, 0.5}
\definecolor{mint}{rgb}{0.74, 0.99, 0.79}
\definecolor{coral}{rgb}{1.0, 0.5, 0.31}

\definecolor{PeachPuff}{rgb}{1.0, 0.85, 0.73}
\definecolor{Beige}{rgb}{0.96, 0.96, 0.86}
\definecolor{LightSalmon}{rgb}{1.0, 0.63, 0.48}
\definecolor{Ivory}{rgb}{1.0, 1.0, 0.94}
\definecolor{MintCream}{rgb}{0.96, 1.0, 0.98}

\definecolor{Red}{rgb}{1.0, 0.0, 0.0}
\definecolor{Green}{rgb}{0.0, 0.5, 0.0}
\definecolor{Blue}{rgb}{0.0, 0.0, 1.0}
\definecolor{Orange}{rgb}{1.0, 0.55, 0.0}
\definecolor{Purple}{rgb}{0.5, 0.0, 0.5}
\definecolor{Goldenrod}{rgb}{0.85, 0.65, 0.13}
\definecolor{BurntOrange}{HTML}{CC5500}
\definecolor{Cyan}{rgb}{0.0, 1.0, 1.0}
\definecolor{Maroon}{rgb}{0.5, 0.0, 0.0}  % Standard maroon RGB

\definecolor{textblue}{RGB}{25,25,112}      % MidnightBlue variant
\definecolor{textred}{RGB}{139,0,0}         % DarkRed variant
\definecolor{Pink}{RGB}{255,192,203}  % defines 'Pink' as RGB(255,192,203)
\definecolor{softgray}{RGB}{220,220,220}
\definecolor{softblue}{RGB}{230,245,255}
\definecolor{softred}{RGB}{255,235,238}
\definecolor{framegray}{RGB}{100,100,100}
\definecolor{frameblue}{RGB}{100,149,237}   % CornflowerBlue
\definecolor{framered}{RGB}{220,20,60}      % Crimson

\definecolor{LightGreen}{rgb}{0.88, 1.0, 0.88}
\definecolor{LightPink}{rgb}{1.0, 0.9, 0.9}
\definecolor{LemonChiffon}{rgb}{1.0, 0.98, 0.8}

\definecolor{Brown}{rgb}{0.65, 0.16, 0.16}
\definecolor{LightSalmon}{rgb}{1.0, 0.63, 0.48}  % If not defined earlier
\definecolor{Gray}{rgb}{0.5, 0.5, 0.5}           % 60% gray approximation
\definecolor{DarkGreen}{rgb}{0.0, 0.39, 0.0}

\newcommand{\cross}{\ding{55}} % Cross mark

\title{ChildGuard: A Specialized Dataset for Combatting Child-Targeted Hate Speech}

\author{
\textbf{Gautam Siddharth Kashyap}\textsuperscript{1},
\textbf{Mohammad Anas Azeez}\textsuperscript{2},
\textbf{Rafiq Ali}\textsuperscript{3},
\textbf{Zohaib Hasan Siddiqui}\textsuperscript{4}, \\
\textbf{Jiechao Gao}\textsuperscript{5}\thanks{Corresponding Author: jiechao@stanford.edu},
\textbf{Usman Naseem}\textsuperscript{1} \\
\textsuperscript{1}Macquarie University, Sydney, Australia \\
\textsuperscript{2}MBZUAI, Abu Dhabi, United Arab Emirates \\
\textsuperscript{3}DSEU, New Delhi, India \\
\textsuperscript{4}Department of Information and Computer Science, KFUPM, Dhahran, Saudi Arabia \\
\textsuperscript{5}Stanford University, California, USA \\
}

\begin{document}
\maketitle

\begin{abstract}
Mental health industry faces growing concerns regarding hate speech directed at children's on social media, as exposure to such content can contribute to adverse psychological outcomes during critical stages of development. Current hate speech datasets and detection systems provide limited support for child-focused applications because they are primarily designed for adults and lack dedicated representations of age-specific characteristics associated with hate speech directed at children's. To address this gap, we introduce \textbf{ChildGuard}\footnote{\scriptsize{\url{https://github.com/gskgautam/ChildGuard}}}, a large-scale English dataset for child-targeted hate speech containing 351,877 annotated instances collected from X (formerly Twitter), Reddit, and YouTube. The dataset covers three age groups such as younger children's (under 11), pre-teens (11-12), and teens (13-17). \textbf{ChildGuard} contains two subsets such as a contextual subset (157K) and a lexical subset (194K). Evaluation using recent transformer-based models and LLMs achieves a best Macro-F1 of 82.07\%, decreasing to 79.41\%, 79.24\%, 76.04\%, and 74.88\% on younger children's, contextual, implicit hate, and cross-subset settings, respectively.
\end{abstract}

\section{Introduction}
\label{Introduction}

\begin{center}
\small
\colorbox{red!10}{\parbox{0.92\linewidth}{\centering
\textit{You're too fat to dance. Nobody wants to see you.}
}}

\vspace{0.15cm}

\colorbox{blue!10}{\parbox{0.92\linewidth}{\centering
\textit{Go back to your country.}
}}

\vspace{0.15cm}

\colorbox{orange!10}{\parbox{0.92\linewidth}{\centering
\textit{You sound like a baby. Kill yourself already.}
}}
\end{center}

\begin{figure}[t]
\vspace{-0.3cm}
\centering
\includegraphics[width=\linewidth]{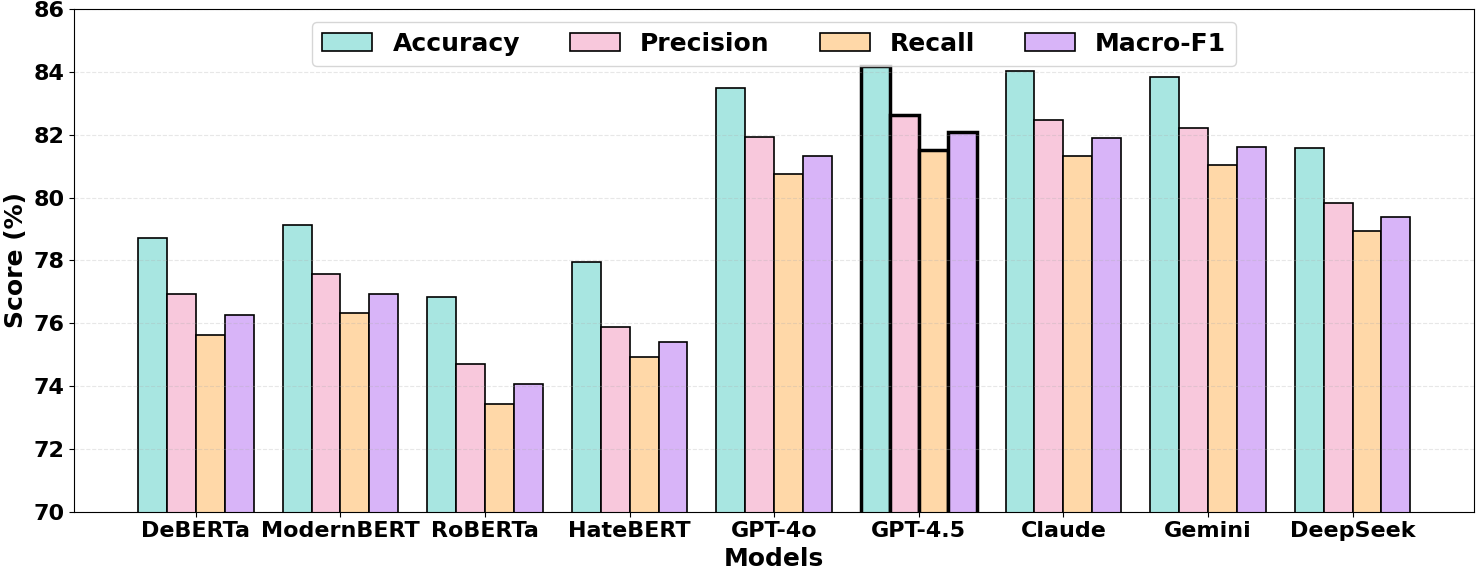}
\vspace{-0.8cm}
\caption{
Illustration of the performance of prior SOTA systems on child-targeted hate speech. Despite a best Macro-F1 of 82.07\% on \textbf{ChildGuard}, age-specific and contextual characteristics remain challenging, motivating the need for \textbf{ChildGuard} in mental health industry applications (see \S\ref{Downstream Analysis}).
}
\label{fig:motivation}
\vspace{-0.3cm}
\end{figure}

\begin{figure*}[t]
\vspace{-0.3cm}
\centering
\includegraphics[width=15cm]{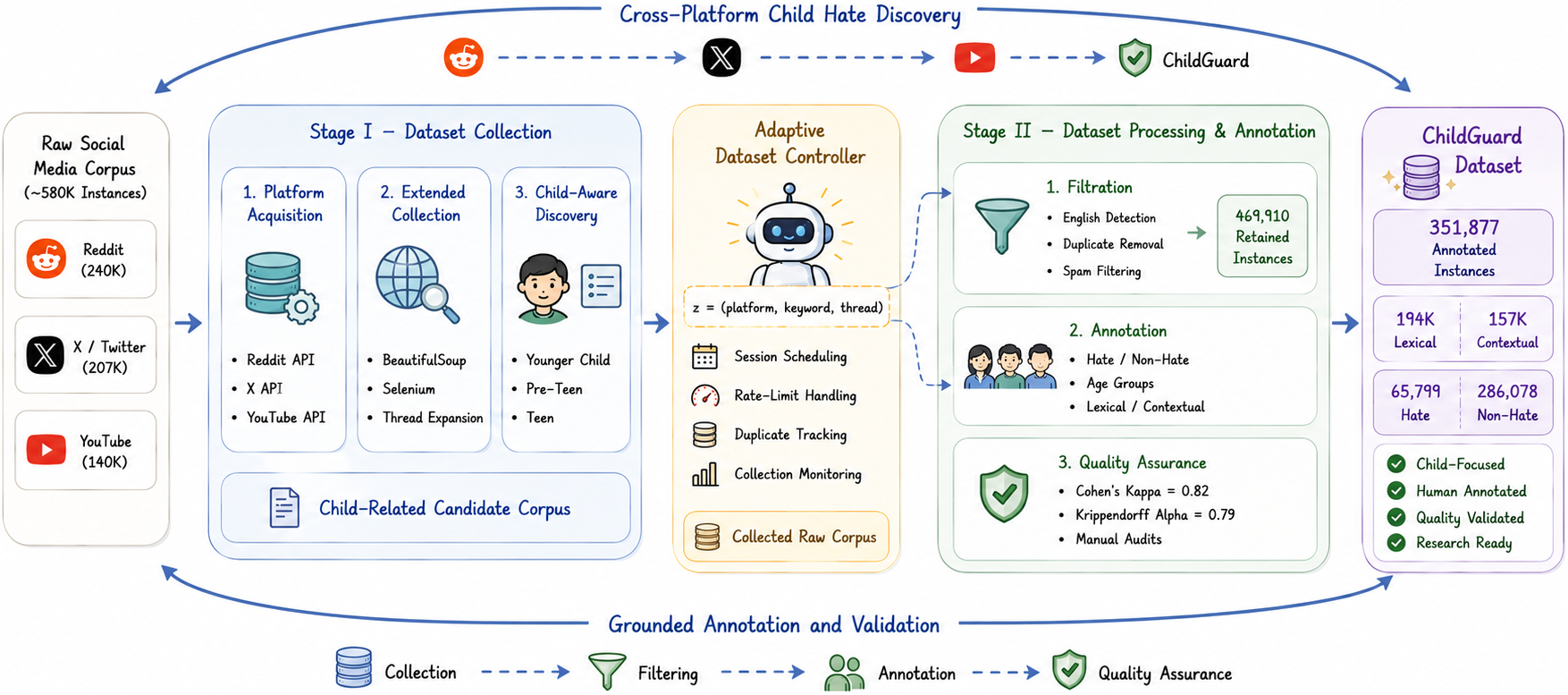}
\vspace{-0.3cm}
\caption{
Construction pipeline of \textbf{ChildGuard}. Child-related content was collected from Reddit, X, and YouTube, followed by filtration, anonymity processing, and manual annotation. Instances were labeled for hate speech and age-group categories corresponding to younger children's, pre-teens, and teens, and organized into lexical and contextual subsets. The resulting dataset contains 351,877 annotated instances to support child-focused hate speech analysis in mental health industry applications.
}
\label{fig:arch}
\vspace{-0.3cm}
\end{figure*}

Mental health industry faces growing concerns regarding hate speech directed at children's on social media, where increasing participation in online platforms exposes children's to hostility targeting their age, appearance, ethnicity, nationality, and disabilities \cite{kurki2026hyper}. Since children's often possess limited emotional resilience and cognitive maturity, such exposure can contribute to emotional distress, anxiety, social isolation, reduced self-esteem, and other adverse psychological outcomes during critical stages of development \cite{cho2026becoming, aseeri2026reality}. However, current hate speech datasets and detection systems (e.g., \citealt{fesaghandis2026multilingual, albladi2025hate}) remain largely adult-centric and provide limited support for child-focused applications. \textit{Most datasets model the relationship between textual content $X$ and hate speech labels $Y \in \{0,1\}$, effectively learning $P(Y|X)$ without accounting for age-specific characteristics, limiting their effectiveness for child-targeted hate speech detection} (see Figure \ref{fig:motivation}).

To address this gap, we introduce \textbf{ChildGuard}, a large-scale English dataset for child-targeted hate speech containing 351,877 annotated instances collected from X\footnote{\scriptsize{\url{https://twitter.com}}} (formerly Twitter), Reddit\footnote{\scriptsize{\url{https://reddit.com}}}, and YouTube\footnote{\scriptsize{\url{https://youtube.com}}}. \textbf{ChildGuard} includes age-group annotations $A \in \{\text{younger child}, \text{pre-teen}, \text{teen}\}$ corresponding to younger children's (under 11), pre-teens (11-12), and teens (13-17), reflecting established developmental differences across childhood \cite{panchanadikar2026beyond}. The dataset further contains a contextual subset (157K) focused on discourse-level characteristics and a lexical subset (194K) focused on vocabulary and sentiment patterns, supporting the study of relationships among textual content $X$, age-group information $A$, and hate speech labels $Y$ beyond traditional formulations that model only $P(Y|X)$ (e.g., \citealt{fesaghandis2026multilingual, albladi2025hate}). In summary, the main contribution of this work are as follows: 

\begin{itemize}
\vspace{-0.3cm}
    \item We introduce \textbf{ChildGuard}, a large-scale English dataset for child-targeted hate speech containing 351,877 annotated instances from X, Reddit, and YouTube.
\vspace{-0.3cm}
    \item Empirically, the best-performing model achieves 82.07\% Macro-F1, decreasing to 79.41\%, 79.24\%, 76.04\%, and 74.88\% on younger children's, contextual, implicit hate, and cross-subset settings, respectively.
\end{itemize}

\section{Related Work}
\label{related_work}

Mental health industry faces growing concerns regarding hate speech directed at children's as discussed in \S\ref{Introduction}. Prior works have produced datasets such as Hate Offensive \cite{davidson2017automated}, Slur Corpus \cite{kurrek2020towards}, HateXplain \cite{mathew2021hatexplain}, ETHOS \cite{mollas2022ethos}, Gab Hate Corpus \cite{kennedy2022introducing}, Measuring Hate Speech \cite{sachdeva2022measuring}, Curated Hate Speech \cite{mody2023curated}, Hate Comments \cite{gupta2023hateful}, Context Toxicity \cite{pavlopoulos2020toxicity}. Global School-Based Student Health Survey (GSHS)\footnote{\scriptsize{\url{https://www.kaggle.com/datasets/leomartinelli/bullying-in-schools}}} and BullyDetect\footnote{\scriptsize{\url{https://www.kaggle.com/datasets/laptype/wifi-bullydetect}}}, which primarily model the relationship between textual content $X$ and hate speech labels $Y \in \{0,1\}$ across demographic categories including race, religion, ethnicity, nationality, gender, and political affiliation. Recent works have further improved hate speech datasets and hate speech detection systems through transformer-based architectures \cite{fesaghandis2026multilingual} and Large Language Models (LLMs) \cite{albladi2025hate}. However, these works remain largely adult-centric and rarely incorporate age-group information $A$, limiting their effectiveness in child-focused applications. Therefore, datasets specifically designed for child-targeted hate speech remain scarce, motivating the need for \textbf{ChildGuard} (see Table \ref{tab:child_dataset_comparison}).

\begin{table}[t]
\vspace{-0.6cm}
\centering
\scriptsize
\renewcommand{\arraystretch}{1.05}
\setlength{\tabcolsep}{3pt}

\begin{tabular}{lcccc}
\toprule
\textbf{Dataset} &
\cellcolor{coral!20}\textbf{Child} &
\cellcolor{MintCream!195}\textbf{Age} &
\cellcolor{Beige!85}\textbf{Context}
\textbf{Size} \\
\midrule
Hate Offensive & \cross & \cross & \cross & 24,783 \\
Slur Corpus & \cross & \cross & \cross & 39,960 \\
HateXplain & \cross & \cross & \checkmark & 20,109 \\
ETHOS & \cross & \cross & \cross & 1,441 \\
Gab Hate Corpus & \cross & \cross & \cross & 27,434 \\
Measuring Hate Speech & \cross & \cross & \cross & 68,597 \\
Curated Hate Speech & \cross & \cross & \cross & 560,385 \\
HateComments & \cross & \cross & \cross & 2,070 \\
Context Toxicity & \cross & \cross & \checkmark & 19,842 \\
GSHS & \checkmark & \cross & \cross & -- \\
BullyDetect & \checkmark & \cross & \cross & -- \\
\midrule
\textbf{ChildGuard} &
\cellcolor{coral!20}\checkmark &
\cellcolor{MintCream!195}\checkmark &
\cellcolor{Beige!85}\checkmark &
\textbf{351,877} \\
\bottomrule
\end{tabular}

\vspace{-0.3cm}
\caption{Comparison of datasets relevant to child-targeted hate speech for the mental health industry applications, highlighting that \textbf{ChildGuard} is the only dataset combining child focus and age-specific characteristics.}
\label{tab:child_dataset_comparison}
\vspace{-0.3cm}
\end{table}

\section{Overview of the Dataset}
\label{Dataset Overview}

\textbf{ChildGuard} is a large-scale dataset for child-targeted hate speech developed to support child-focused applications in the mental health industry applications as discussed in \S\ref{Introduction} (see Figure \ref{fig:arch}).  

\subsection{Dataset Collection}
\label{Dataset Construction}

To support the study of child-targeted hate speech in the mental health industry, we independently collected approximately 580,000 raw text instances from three major online platforms where children's actively participate such as Reddit (240K), X (207K), and YouTube (140K). These platforms were selected because of their widespread use among children's, large volumes of public discussions, and suitability for large-scale analysis. No pre-annotated datasets were reused during data collection. Data acquisition was conducted using publicly available platform interfaces, including PRAW\footnote{\scriptsize{\url{https://praw.readthedocs.io/}}} for Reddit, Tweepy\footnote{\scriptsize{\url{https://docs.tweepy.org/en/stable/}}} for X, and the YouTube Data API v3\footnote{\scriptsize{\url{https://developers.google.com/youtube/v3}}} for YouTube, where X queries were executed using keyword-based retrieval and public conversation metadata. In situations where standard API responses provided incomplete access to discussion threads, such as omitted replies within older YouTube comments or restrictions on Reddit thread depth, we developed platform-specific collection scripts using BeautifulSoup\footnote{\scriptsize{\url{https://www.crummy.com/software/BeautifulSoup/bs4/doc/}}} and Selenium\footnote{\scriptsize{\url{https://www.selenium.dev/documentation/}}}. Requests and BeautifulSoup were used to retrieve static HTML content and associated metadata, while Selenium was used to load dynamic discussion threads on YouTube that were not consistently available through standard API responses.

\begin{table}[t]
\vspace{-0.3cm}
\centering
\scriptsize
\renewcommand{\arraystretch}{1.05}
\setlength{\tabcolsep}{2pt}

\begin{tabular}{lccc}
\toprule
\textbf{Keyword} &
\cellcolor{green!15}\textbf{Reddit} &
\cellcolor{violet!15}\textbf{X} &
\cellcolor{cyan!15}\textbf{YouTube} \\
\midrule
kid & 18,420 & 15,670 & 11,240 \\
baby & 11,310 & 9,830 & 7,640 \\
school & 6,370 & 5,420 & 4,270 \\
brat & 5,110 & 4,060 & 2,970 \\
teenager & 12,540 & 10,750 & 8,210 \\
go play with toys & 3,820 & 3,140 & 2,200 \\
little one & 3,120 & 2,650 & 1,780 \\
child & 13,410 & 11,250 & 8,450 \\
children & 9,680 & 8,110 & 6,150 \\
minor & 4,010 & 3,420 & 2,560 \\
schoolgirl & 4,730 & 3,950 & 2,880 \\
toddler & 4,480 & 3,700 & 2,900 \\
diaper & 1,740 & 1,380 & 930 \\
playschool & 1,520 & 1,230 & 790 \\
playground & 3,540 & 3,000 & 2,300 \\
preschool & 2,320 & 1,950 & 1,410 \\
babyface & 1,360 & 1,120 & 700 \\
kiddie & 1,860 & 1,560 & 1,120 \\
childish & 3,820 & 3,200 & 2,550 \\
nursery & 1,510 & 1,270 & 1,010 \\
\midrule
\textbf{Total} &
\cellcolor{green!15}\textbf{240,840} &
\cellcolor{violet!15}\textbf{207,680} &
\cellcolor{cyan!15}\textbf{143,770} \\
\bottomrule
\end{tabular}

\vspace{-0.3cm}
\caption{Keyword-based collection statistics across Reddit, X, and YouTube for child-targeted hate speech.}
\label{tab:raw_keyword_counts}
\vspace{-0.3cm}
\end{table}

API-based collection followed the developer policies of Reddit, X, and YouTube, while web-based collection scripts followed publicly available platform policies and \texttt{robots.txt}\footnote{\scriptsize{\url{https://developers.google.com/search/docs/crawling-indexing/robots/intro}}} guidelines, including those of Reddit\footnote{\scriptsize{\url{https://redditinc.com/policies}}}, X\footnote{\scriptsize{\url{https://help.x.com/en/rules-and-policies}}}, and YouTube\footnote{\scriptsize{\url{https://www.youtube.com/howyoutubeworks/our-policies/}}}. Request intervals of 1.5 second-3 seconds were introduced between consecutive queries, and commonly used browser signatures, including \texttt{Chrome/114}, \texttt{Safari/14}, \texttt{Firefox/89}, and \texttt{Mobile Safari/14}, were rotated across collection sessions. Each session randomly selected one browser signature and typically lasted 15 minutes-30 minutes, collecting approximately 1,000 posts-1,500 posts or comments depending on the platform. Across all platforms, approximately 600 collection sessions were executed, including retries associated with timeouts and rate limits. Variations in the number of retrieved instances arose from duplicate removal, platform-specific filtering, incomplete discussion threads, and keyword-based relevance filtering during corpus construction.

\begin{figure*}[t!]
\vspace{-0.3cm}
    \centering
    \begin{subfigure}{0.32\textwidth}
        \centering
        \includegraphics[width=\linewidth]{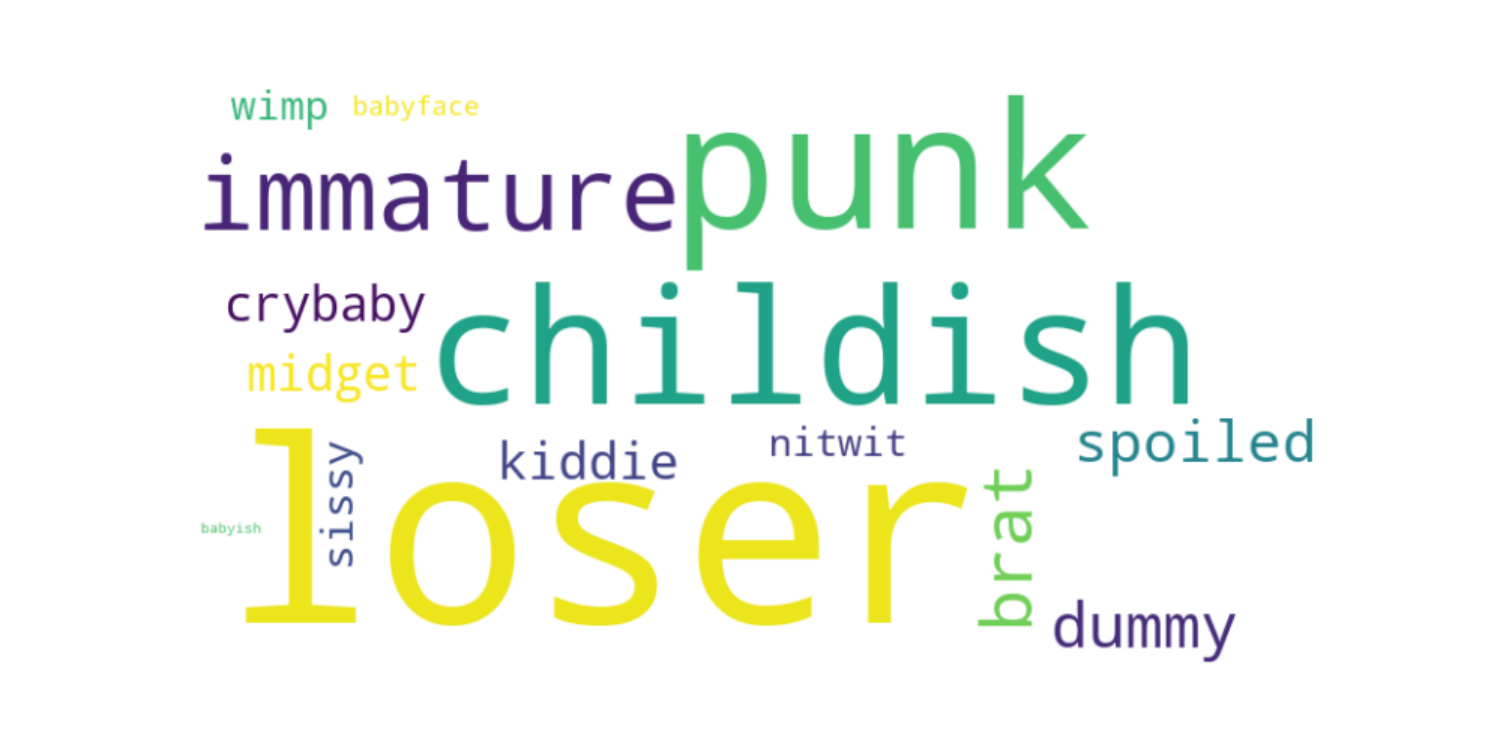}
        \caption{Hate Speech}
    \end{subfigure}%
    %\hspace{0.05\textwidth} 
    \begin{subfigure}{0.32\textwidth}
        \centering
        \includegraphics[width=\linewidth]{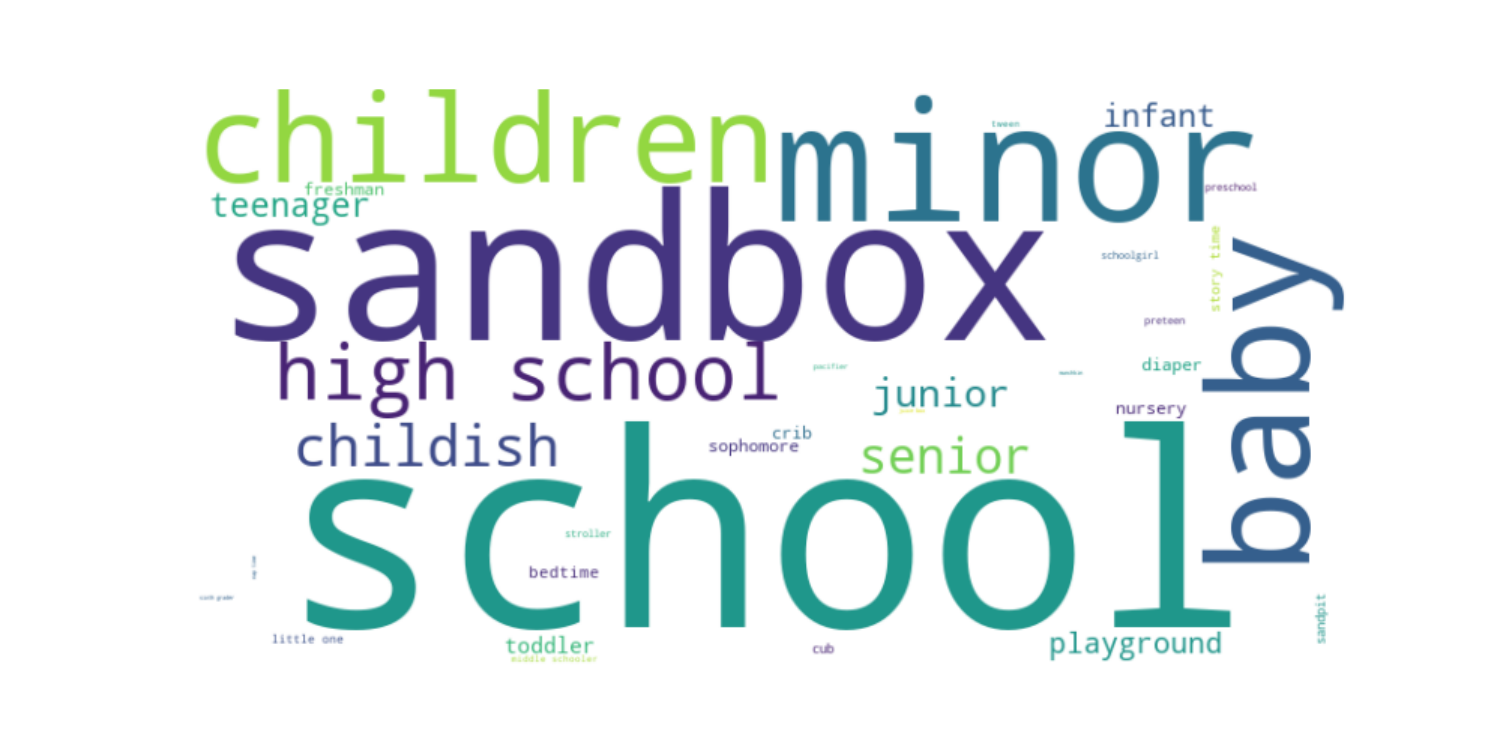}
        \caption{Age Groups}
    \end{subfigure}%
    %\hspace{0.05\textwidth} 
    \begin{subfigure}{0.32\textwidth}
        \centering
        \includegraphics[width=\linewidth]{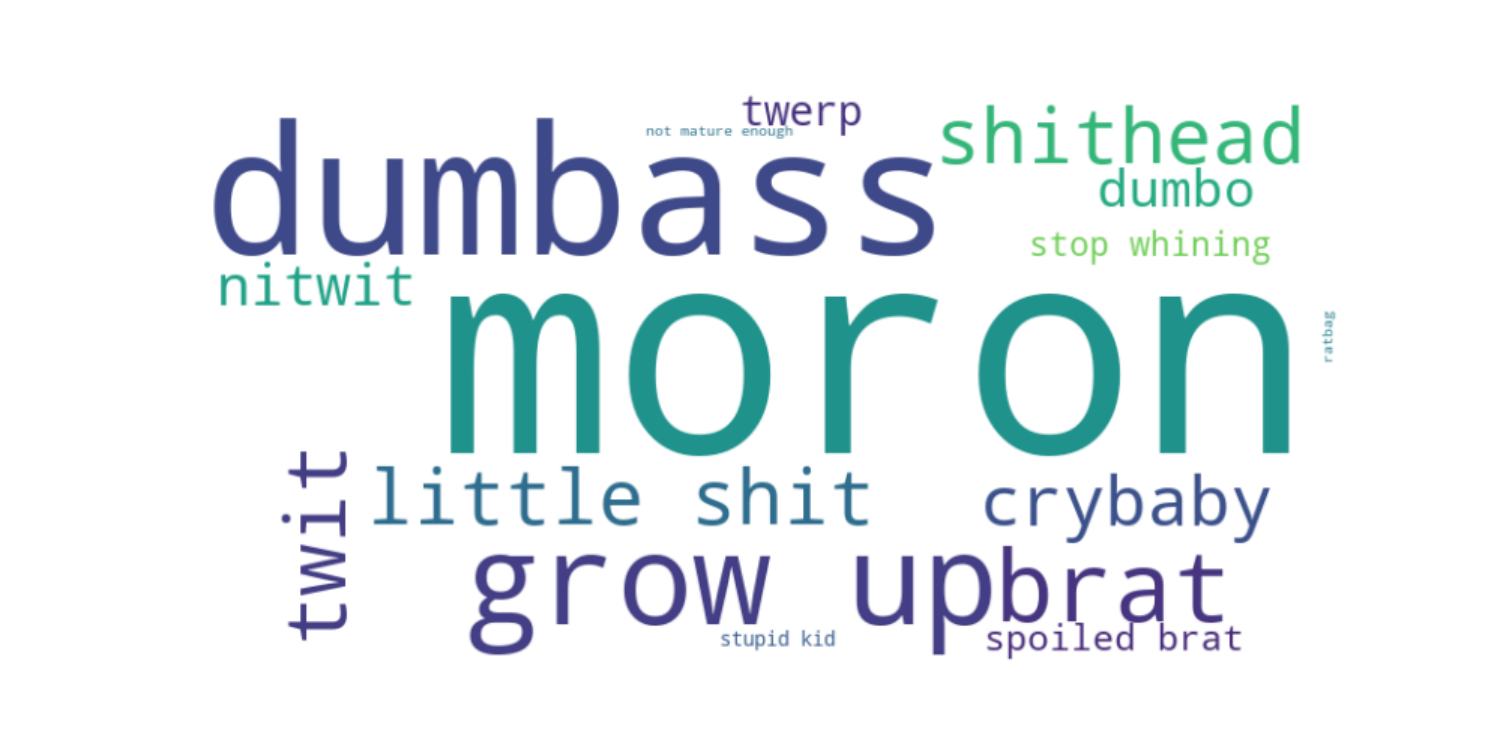}
        \caption{Lexical vs. Contextual}
    \end{subfigure}
    \vspace{-0.3cm}
    \caption{Illustration of linguistic characteristics associated with child-targeted hate speech in \textbf{ChildGuard}. Word clouds highlight representative terms across (a) hate speech annotations, (b) age-group categories corresponding to younger children's, pre-teens, and teens, and (c) lexical and contextual subsets designed to capture vocabulary-level and discourse-level characteristics relevant to child-focused applications.}
    \label{word}
\vspace{-0.3cm}
\end{figure*}

The collection process was conducted over approximately three weeks by four authors working in parallel. All collected content was publicly accessible and did not require login credentials, paid subscriptions, authentication bypass, or access-control circumvention. To identify content relevant to children's, we constructed a lexicon of child-related keywords and phrases spanning younger children's, pre-teens, and teens through an iterative manual review of platform content (see Table~\ref{tab:raw_keyword_counts}). The same keyword set was applied across Reddit, X, and YouTube to maintain consistent coverage of age-specific characteristics associated with child-targeted hate speech.

\subsection{Dataset Filtration}
\label{Dataset Filtration}

Following the collection process described in \S\ref{Dataset Construction}, we applied a multi-stage filtration pipeline to remove irrelevant content, spam, non-English text, and duplicate entries prior to annotation. Language identification was performed using the \texttt{langdetect}\footnote{\scriptsize{\url{https://pypi.org/project/langdetect/}}} library, retaining only English-language instances, while duplicate removal relied on SHA-256\footnote{\scriptsize{\url{https://csrc.nist.gov/publications/detail/fips/180/4/final}}} hash comparisons of normalized text strings. Spam filtering used rule-based criteria, including excessive phrase repetition (more than three occurrences), abnormal punctuation patterns, and commonly occurring spam expressions such as ``click here'' and ``subscribe now''. Instances containing fewer than 10 tokens or content dominated by URLs or emojis ($>70\%$) were also removed. All filtration stages were implemented through automated Python scripts without manual intervention. After filtration, 469,910 instances were retained for annotation, including 179,993 instances from Reddit, 169,958 from X, and 119,959 from YouTube (see Table~\ref{tab:filterkeyword_platform_counts}).

\begin{table}[t]
\vspace{-0.3cm}
\centering
\scriptsize
\renewcommand{\arraystretch}{1.05}
\setlength{\tabcolsep}{2pt}

\begin{tabular}{lccc}
\toprule
\textbf{Keyword} &
\cellcolor{green!15}\textbf{Reddit} &
\cellcolor{violet!15}\textbf{X} &
\cellcolor{cyan!15}\textbf{YouTube} \\
\midrule
kid & 15,287 & 14,463 & 10,157 \\
baby & 9,987 & 9,418 & 7,185 \\
school & 5,089 & 4,903 & 3,814 \\
brat & 4,598 & 4,053 & 2,689 \\
teenager & 11,175 & 10,238 & 6,993 \\
go play with toys & 3,312 & 2,849 & 1,707 \\
little one & 2,709 & 2,391 & 1,498 \\
child & 12,994 & 11,987 & 8,408 \\
children & 9,274 & 8,612 & 6,089 \\
minor & 3,790 & 3,405 & 2,491 \\
schoolgirl & 4,293 & 3,886 & 2,103 \\
toddler & 4,085 & 3,693 & 1,996 \\
diaper & 1,693 & 1,397 & 848 \\
playschool & 1,498 & 1,242 & 715 \\
playground & 3,485 & 3,199 & 2,192 \\
preschool & 2,297 & 1,997 & 1,305 \\
babyface & 1,295 & 1,146 & 796 \\
kiddie & 1,792 & 1,601 & 1,098 \\
childish & 3,589 & 3,187 & 2,087 \\
nursery & 1,393 & 1,198 & 847 \\
\midrule
\textbf{Total} &
\cellcolor{green!15}\textbf{179,993} &
\cellcolor{violet!15}\textbf{169,958} &
\cellcolor{cyan!15}\textbf{119,959} \\
\bottomrule
\end{tabular}

\vspace{-0.3cm}
\caption{Keyword-based filtration statistics across Reddit, X, and YouTube for child-targeted hate speech.}
\label{tab:filterkeyword_platform_counts}
\vspace{-0.3cm}
\end{table}

\begin{table}[t]
\vspace{-0.3cm}
\scriptsize
\centering
\renewcommand{\arraystretch}{1.0}
\setlength{\tabcolsep}{4pt}

\begin{tabularx}{\columnwidth}{>{\raggedright\arraybackslash}X *{3}{>{\centering\arraybackslash}X}}
\toprule
\textbf{Statistic} &
\cellcolor{lime!20}\textbf{Contextual} &
\cellcolor{magenta!15}\textbf{Lexical} &
\cellcolor{brown!15}\textbf{ChildGuard} \\
\midrule
Unique Words (Vocabulary) & 22,890 & 28,764 & 35,412 \\
Average Text Length (Tokens) & 22.7 & 19.1 & 21.2 \\
Minimum Text Length Threshold & 10 & 10 & 10 \\
Entries Containing Emojis & 35,700 & 30,100 & 65,500 \\
Number of Dependency Trees & 194,597 & 157,280 & 351,877 \\
Proportion of Explicit Hate & 24\% & 26\% & 25\% \\
Proportion of Implicit Hate & 14\% & 12\% & 13\% \\
Platform Coverage & X, Reddit, YouTube & X, Reddit, YouTube & X, Reddit, YouTube 
\\
\bottomrule
\end{tabularx}

\vspace{-0.3cm}
\caption{Dataset statistics for the contextual subset, lexical subset, and \textbf{ChildGuard}.}
\label{tab:childguard_additional_stats_revised}
\vspace{-0.2cm}
\end{table}

\begin{table}[t]
\centering
\scriptsize
\renewcommand{\arraystretch}{1.0}
\setlength{\tabcolsep}{4pt}

\begin{tabularx}{\columnwidth}{>{\raggedright\arraybackslash}X *{4}{>{\centering\arraybackslash}X}}
\toprule
\textbf{Data Split} &
\cellcolor{blue!15}\textbf{Total} &
\cellcolor{red!15}\textbf{Train} &
\cellcolor{yellow!20}\textbf{Val} &
\cellcolor{gray!20}\textbf{Test} \\
\midrule

\textbf{ChildGuard} & 351,877 & 246,314 & 52,781 & 52,782 \\
\textbf{Lexical} & 157,280 & 110,096 & 23,592 & 23,592 \\
\textbf{Contextual} & 194,597 & 136,120 & 29,239 & 29,238 \\

\midrule

\textbf{Teens} & 11,479 & 8,035 & 1,721 & 1,723 \\
\textbf{Pre-teens} & 5,950 & 4,165 & 893 & 892 \\
\textbf{Younger Children} & 34,300 & 23,010 & 5,145 & 6,145 \\

\bottomrule
\end{tabularx}

\vspace{-0.3cm}
\caption{Sample allocation for \textbf{ChildGuard} and its contextual and lexical subsets.}
\label{tab:sample_allocation}
\vspace{-0.3cm}
\end{table}

\subsection{Annotations}
\label{Annotations}

Following the filtration process described in \S\ref{Dataset Filtration}, all retained instances were manually annotated by three authors with prior experience in hate speech research and linguistic annotation. Consistent with the objectives of \textbf{ChildGuard}, each instance was assigned a hate speech label $Y \in \{0,1\}$ and an age-group label $A \in \{\text{younger child}, \text{pre-teen}, \text{teen}\}$. Instances were labeled as hateful ($Y=1$) when they contained derogatory, hostile, threatening, or demeaning language directed at children's. Examples included (see Figure \ref{word}) terms such as \textit{``worthless brat''}, \textit{``ugly kid''}, \textit{``idiot baby''}, \textit{``babyface''}, \textit{``kiddie''}, and \textit{``childish''}. Terms such as \textit{``toddler''}, \textit{``diaper''}, \textit{``nursery''}, \textit{``preschool''}, and \textit{``playschool''} were associated with younger children's (under 11), \textit{``children''}, \textit{``minor''}, \textit{``schoolgirl''}, and \textit{``playground''} informed pre-teen annotations (11-12), and terms such as \textit{``teenager''}, \textit{``school''}, and \textit{``high school''} informed teen annotations (13-17). Instances for which annotations could not be reliably determined were excluded from the final corpus.

The subset category $S \in \{\text{lexical}, \text{contextual}\}$ distinguishes between hate speech expressed through vocabulary and sentiment patterns and hate speech requiring broader discourse-level interpretation. The lexical subset captures expressions conveyed through offensive terms and direct insults, while the contextual subset captures conversational context, indirect references, sarcasm, and demeaning discourse patterns. Each instance was independently annotated by two authors, with disagreements reviewed by a third author. Prior to large-scale annotation, a pilot study containing 1,000 instances was conducted to refine annotation guidelines, followed by annotation in batches of 10,000 instances-15,000 instances over approximately four weeks. Inter-annotator agreement measured using Cohen's Kappa and Krippendorff's Alpha yielded scores of 0.82 and 0.79, respectively. Automated validation scripts and manual audits were subsequently used to identify missing labels, annotation inconsistencies, and unreliable instances, which were removed from the corpus. After final validation, \textbf{ChildGuard} contains 351,877 annotated instances, comprising a lexical subset of 194K instances and a contextual subset of 157K instances. Across the complete corpus, 65,799 instances are annotated as hate speech and 286,078 as non-hate speech. Additional descriptive statistics are presented in Tables~\ref{tab:childguard_additional_stats_revised} and \ref{tab:sample_allocation}.

\section{Experimental Setup}

\paragraph{Baselines.}
To assess the difficulty of child-targeted hate speech detection, we conduct a downstream evaluation using recent transformer-based models and LLMs selected for their strong performance on text classification, hate speech detection, and language understanding tasks. Transformer baselines include DeBERTa-v3-large (DeBERTa)\footnote{\scriptsize{\url{https://huggingface.co/microsoft/deberta-v3-large}}}, ModernBERT\footnote{\scriptsize{\url{https://huggingface.co/answerdotai/ModernBERT-large}}}, RoBERTa-large (RoBERTa)\footnote{\scriptsize{\url{https://huggingface.co/FacebookAI/roberta-large}}}, and HateBERT\footnote{\scriptsize{\url{https://huggingface.co/GroNLP/hateBERT}}}. We further evaluate GPT-4o\footnote{\scriptsize{\url{https://platform.openai.com/docs/models/gpt-4o}}}, GPT-4.5\footnote{\scriptsize{\url{https://openai.com/index/introducing-gpt-4-5/}}}, Claude-3.7 Sonnet (Claude)\footnote{\scriptsize{\url{https://www.anthropic.com/news/claude-3-7-sonnet}}}, Gemini-2.5 Pro (Gemini)\footnote{\scriptsize{\url{https://deepmind.google/technologies/gemini/pro/}}}, and DeepSeek-V3 (DeepSeek)\footnote{\scriptsize{\url{https://huggingface.co/deepseek-ai/DeepSeek-V3-0324}}}. These models are evaluated on \textbf{ChildGuard} across age groups and lexical and contextual subsets. Higher values indicate better performance.

\vspace{-0.2cm}
\paragraph{Dataset and Preprocessing.}
Experiments were conducted on \textbf{ChildGuard}, comprising 351,877 annotated instances from Reddit, X, and YouTube. The dataset includes age-group annotations covering younger children's, pre-teens, and teens with lexical and contextual subsets. Data splits follow Table~\ref{tab:sample_allocation}. Transformer models were tokenized using model-specific HuggingFace tokenizers and truncated or padded to 128 tokens, while LLMs were evaluated using instruction-based prompts with label-constrained outputs.

\vspace{-0.2cm}
\paragraph{Training Setup.}
Transformer models were fine-tuned using AdamW ($2 \times 10^{-5}$ learning rate, batch size 32) for up to 10 epochs with early stopping (patience = 3). LLMs were evaluated in zero-shot settings using instruction-based prompts and label-constrained outputs. All experiments were conducted on a single NVIDIA A100 GPU (40GB VRAM) using PyTorch 2.1 and Ubuntu 22.04.

\vspace{-0.2cm}
\paragraph{Evaluation Metrics.}
Performance was evaluated on the held-out test set using accuracy, precision, recall, and macro-averaged F1 score. Results are reported on the complete \textbf{ChildGuard} dataset as well as across age-group categories and lexical and contextual subsets, assessing model performance on age-specific and discourse-level variations of child-targeted hate speech. All values are in \%.

\begin{figure*}[t!]
\vspace{-0.3cm}
    \centering
    \begin{subfigure}{0.32\textwidth}
        \centering
        \includegraphics[width=\linewidth]{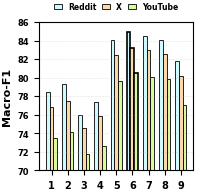}
        \caption{Platform-wise Performance}
    \end{subfigure}%
    \begin{subfigure}{0.32\textwidth}
        \centering
        \includegraphics[width=\linewidth]{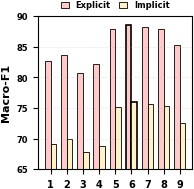}
        \caption{Hate Type Analysis}
    \end{subfigure}%
    \begin{subfigure}{0.32\textwidth}
        \centering
        \includegraphics[width=\linewidth]{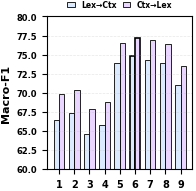}
        \caption{Cross-Subset Generalization}
    \end{subfigure}
    \vspace{-0.3cm}
    \caption{Empirical analysis of model performance on \textbf{ChildGuard}. (a) Performance across Reddit, X, and YouTube demonstrates platform-specific variation, with lower performance on YouTube. (b) Performance across explicit and implicit hate speech shows that implicit hate remains substantially more difficult to identify. (c) Cross-subset evaluation between lexical and contextual subsets reveals notable performance degradation, highlighting limited generalization across vocabulary-level and discourse-level hate characteristics. Model IDs: 1-DeBERTa, 2-ModernBERT, 3-RoBERTa, 4-HateBERT, 5-GPT-4o, 6-GPT-4.5, 7-Claude, 8-Gemini, 9-DeepSeek.}
    \label{fig:downstream_analysis}
\vspace{-0.3cm}
\end{figure*}

\begin{comment}

\begin{table}[t]
\vspace{-0.3cm}
\centering
\scriptsize
\renewcommand{\arraystretch}{1.0}
\setlength{\tabcolsep}{4pt}

\begin{tabularx}{\columnwidth}{>{\raggedright\arraybackslash}X *{4}{>{\centering\arraybackslash}X}}
\toprule
\textbf{Model} &
\cellcolor{teal!15}\textbf{Accuracy} &
\cellcolor{pink!15}\textbf{Precision} &
\cellcolor{orange!15}\textbf{Recall} &
\cellcolor{purple!15}\textbf{Macro-F1} \\
\midrule

\textbf{DeBERTa} & 78.72 & 76.94 & 75.61 & 76.27 \\
\textbf{ModernBERT} & 79.11 & 77.58 & 76.32 & 76.94 \\
\textbf{RoBERTa} & 76.83 & 74.71 & 73.44 & 74.07 \\
\textbf{HateBERT} & 77.96 & 75.88 & 74.92 & 75.40 \\

\midrule

\textbf{GPT-4o} & 83.47 & 81.92 & 80.74 & 81.33 \\
\textbf{GPT-4.5} &
\textbf{\cellcolor{teal!15}84.18} &
\textbf{\cellcolor{pink!15}82.64} &
\textbf{\cellcolor{orange!15}81.51} &
\textbf{\cellcolor{purple!15}82.07} \\
\textbf{Claude} & 84.03 & 82.47 & 81.33 & 81.89 \\
\textbf{Gemini} & 83.84 & 82.21 & 81.02 & 81.61 \\
\textbf{DeepSeek} & 81.58 & 79.84 & 78.92 & 79.37 \\

\bottomrule
\end{tabularx}

\vspace{-0.3cm}
\caption{Overall benchmark performance on the \textbf{ChildGuard} test set.}
\label{tab:overall_childguard_results}
\vspace{-0.3cm}
\end{table}
\end{comment}

\begin{table}[t]
\vspace{-0.3cm}
\centering
\scriptsize
\renewcommand{\arraystretch}{1.0}
\setlength{\tabcolsep}{4pt}

\begin{tabularx}{\columnwidth}{>{\raggedright\arraybackslash}X *{3}{>{\centering\arraybackslash}X}}
\toprule
\textbf{Model} &
\cellcolor{cyan!15}\textbf{Teens} &
\cellcolor{lime!15}\textbf{Pre-Teens} &
\cellcolor{rose!15}\textbf{Younger Children} \\
\midrule

\textbf{DeBERTa} & 78.84 & 76.13 & 72.88 \\
\textbf{ModernBERT} & 79.52 & 77.04 & 73.67 \\
\textbf{RoBERTa} & 76.92 & 74.08 & 71.21 \\
\textbf{HateBERT} & 77.71 & 75.36 & 72.04 \\

\midrule

\textbf{GPT-4o} & 84.18 & 81.94 & 78.52 \\
\textbf{GPT-4.5} &
\textbf{\cellcolor{cyan!15}84.97} &
\textbf{\cellcolor{lime!15}82.76} &
\textbf{\cellcolor{rose!15}79.41} \\
\textbf{Claude} & 84.62 & 82.31 & 79.05 \\
\textbf{Gemini} & 84.21 & 81.88 & 78.76 \\
\textbf{DeepSeek} & 82.03 & 79.74 & 76.95 \\

\bottomrule
\end{tabularx}

\vspace{-0.3cm}
\caption{Macro-F1 performance across age-group categories in \textbf{ChildGuard}.}
\label{tab:age_group_results}
\vspace{-0.3cm}
\end{table}

\section{Downstream Analysis}
\label{Downstream Analysis}

\paragraph{Age-Group Performance Analysis.}
Table~\ref{tab:age_group_results} reports Macro-F1 performance across age groups. Performance consistently declines from teens to younger children's across all models, with GPT-4.5 achieving the highest scores in each category. The largest performance drop occurs for younger children's, suggesting that hate speech targeting this group contains more subtle and context-dependent characteristics. These results indicate that age-specific variations remain a major challenge for current hate speech detection systems in mental health industry applications.

\begin{table}[t]
\vspace{-0.3cm}
\centering
\scriptsize
\renewcommand{\arraystretch}{1.0}
\setlength{\tabcolsep}{4pt}

\begin{tabularx}{\columnwidth}{>{\raggedright\arraybackslash}X *{2}{>{\centering\arraybackslash}X}}
\toprule
\textbf{Model} &
\cellcolor{cyan!15}\textbf{Lexical} &
\cellcolor{orange!15}\textbf{Contextual} \\
\midrule

\textbf{DeBERTa} & 78.63 & 73.94 \\
\textbf{ModernBERT} & 79.41 & 74.47 \\
\textbf{RoBERTa} & 76.28 & 71.86 \\
\textbf{HateBERT} & 77.82 & 73.15 \\

\midrule

\textbf{GPT-4o} & 84.37 & 78.29 \\
\textbf{GPT-4.5} &
\textbf{\cellcolor{cyan!15}85.11} &
\textbf{\cellcolor{orange!15}79.24} \\
\textbf{Claude} & 84.82 & 78.97 \\
\textbf{Gemini} & 84.34 & 78.54 \\
\textbf{DeepSeek} & 82.03 & 76.71 \\

\bottomrule
\end{tabularx}

\vspace{-0.3cm}
\caption{Macro-F1 performance across lexical and contextual subsets of \textbf{ChildGuard}.}
\label{tab:lexical_contextual_results}
\vspace{-0.3cm}
\end{table}

\vspace{-0.2cm}
\paragraph{Lexical versus Contextual Performance Analysis.}
Table~\ref{tab:lexical_contextual_results} reports Macro-F1 performance across the lexical and contextual subsets of \textbf{ChildGuard}. All models perform better on the lexical subset, with GPT-4.5 achieving the highest Macro-F1 score of 85.11\%, compared to 79.24\% on the contextual subset. Similar performance gaps are observed across transformer-based models and other LLMs, indicating that hate speech expressed through indirect references, conversational context, and discourse-level characteristics remains substantially more challenging than vocabulary-based hate speech.

\begin{table}[t]
%\vspace{-0.3cm}
\centering
\scriptsize
\renewcommand{\arraystretch}{1.0}
\setlength{\tabcolsep}{4pt}

\begin{tabularx}{\columnwidth}{>{\raggedright\arraybackslash}X >{\centering\arraybackslash}X}
\toprule
\textbf{Error Category} &
\cellcolor{red!15}\textbf{Proportion (\%)} \\
\midrule

Implicit Hate &
\textbf{\cellcolor{red!15}31.4} \\

Context Dependence &
25.8 \\

Age Ambiguity &
18.9 \\

Target Ambiguity &
13.7 \\

Figurative Language &
10.2 \\

\bottomrule
\end{tabularx}

\vspace{-0.3cm}
\caption{Distribution of error categories identified from misclassified instances in \textbf{ChildGuard}.}
\label{tab:error_analysis}
\vspace{-0.3cm}
\end{table}

\vspace{-0.2cm}
\paragraph{Error Analysis.}
Table~\ref{tab:error_analysis} summarizes the major sources of model errors on \textbf{ChildGuard}. Implicit hate accounts for the largest proportion of misclassifications (31.4\%), followed by context dependence (25.8\%) and age ambiguity (18.9\%). Smaller but notable errors arise from target ambiguity (13.7\%) and figurative language (10.2\%). These findings indicate that current models struggle primarily with indirect, context-dependent, and age-specific forms of child-targeted hate speech, highlighting key challenges for mental health industry applications.

\vspace{-0.2cm}
\paragraph{Platform, Hate Type, and Generalization Analysis.}
Figure~\ref{fig:downstream_analysis} provides additional insights into the challenges of child-targeted hate speech detection. Figure~\ref{fig:downstream_analysis}(a) shows that performance is consistently highest on Reddit and lowest on YouTube, with GPT-4.5 achieving Macro-F1 scores of 84.88\%, 83.16\%, and 80.52\%, respectively. Figure~\ref{fig:downstream_analysis}(b) reveals a substantial gap between explicit and implicit hate speech, where GPT-4.5 drops from 88.62\% to 76.04\% Macro-F1, indicating difficulties in identifying indirect hostility and context-dependent abuse. Figure~\ref{fig:downstream_analysis}(c) further demonstrates limited cross-subset generalization, with GPT-4.5 achieving 74.88\% Macro-F1 when transferring from lexical to contextual instances and 77.23\% in the reverse setting.

\section{Conclusion}

Mental health industry applications require reliable detection of child-targeted hate speech due to its impact on psychological well-being and development. We introduce \textbf{ChildGuard}, a large-scale English dataset containing 351,877 annotated instances from Reddit, X, and YouTube, with age-group annotations and lexical and contextual subsets. Empirical results show that current transformer-based models and LLMs continue to struggle, highlighting the need for improved child-focused hate speech detection systems.

\section*{Limitations}
\label{sec:Limitations}

\textbf{ChildGuard} provides a large-scale resource for child-targeted hate speech, it is limited to English-language content from Reddit, X, and YouTube. Age-group annotations are inferred from textual and contextual cues rather than verified demographic information, and some instances remain inherently ambiguous, particularly those involving implicit hate speech and context-dependent expressions. Furthermore, benchmark results are restricted to the evaluated transformer-based models and LLMs, leaving broader architectures and multilingual settings for future work.

\section*{Ethics Statement}
\label{sec:Ethics Statement}

To protect privacy throughout the construction of \textbf{ChildGuard}, all collected instances underwent an anonymity process prior to annotation. Personally Identifiable Information (PII), including names, usernames, and location references, was automatically detected using spaCy's\footnote{\scriptsize{\url{https://spacy.io}}} English Named Entity Recognition (NER) model \texttt{en\_core\_web\_trf}\footnote{\scriptsize{\url{https://spacy.io/models/en}}} together with regular-expression rules designed to identify platform-specific identifiers such as user handles beginning with ``@''. Detected identifiers were replaced with SHA-256 hashed tokens to preserve referential consistency while preventing disclosure of original identities, whereas identifiers that could not be reliably anonymized without affecting interpretation were removed. Only textual content required for hate speech analysis was retained, while metadata such as timestamps and thread identifiers were stored separately at coarse granularity without links to individual identities. No user profiles, private account information, login credentials, or non-public content were collected. All anonymity procedures were applied before annotation and followed the publicly available policies of Reddit, X, and YouTube described in \S\ref{Dataset Construction}.

\bibliography{main}

\begin{thebibliography}{15}
\providecommand{\natexlab}[1]{#1}

\bibitem[{Albladi et~al.(2025)Albladi, Islam, Das, Bigonah, Zhang, Jamshidi, Rahgouy, Raychawdhary, Marghitu, and Seals}]{albladi2025hate}
Aish Albladi, Minarul Islam, Amit Das, Maryam Bigonah, Zheng Zhang, Fatemeh Jamshidi, Mostafa Rahgouy, Nilanjana Raychawdhary, Daniela Marghitu, and Cheryl Seals. 2025.
\newblock Hate speech detection using large language models: A comprehensive review.
\newblock \emph{IEEE Access}, 13:20871--20892.

\bibitem[{Aseeri(2026)}]{aseeri2026reality}
Fatema Khalil~Qasim Aseeri. 2026.
\newblock The reality of cyberbullying among adolescents.
\newblock In \emph{Sustainable Responsible Practices in Technology and Business for Society 5.0: Guideline for Next Generation, Volume 2}, pages 175--183. Springer.

\bibitem[{Cho et~al.(2026)Cho, Hwang, Lam, Lee, and Yip}]{cho2026becoming}
Yeonhee Cho, Daeun Hwang, Ally Lam, Jin~Ha Lee, and Jason~C Yip. 2026.
\newblock Becoming a healthy player: Exploring teen esports players’ perspectives on mental well-being through participatory design.
\newblock In \emph{Proceedings of the 2026 CHI Conference on Human Factors in Computing Systems}, pages 1--19.

\bibitem[{Davidson et~al.(2017)Davidson, Warmsley, Macy, and Weber}]{davidson2017automated}
Thomas Davidson, Dana Warmsley, Michael Macy, and Ingmar Weber. 2017.
\newblock Automated hate speech detection and the problem of offensive language.
\newblock In \emph{Proceedings of the international AAAI conference on web and social media}, volume~11, pages 512--515.

\bibitem[{Fesaghandis and Maity(2026)}]{fesaghandis2026multilingual}
Zahra~Safdari Fesaghandis and Suman~Kalyan Maity. 2026.
\newblock Multilingual hate speech detection and counterspeech generation: A comprehensive survey and practical guide.
\newblock \emph{arXiv preprint arXiv:2603.19279}.

\bibitem[{Gupta et~al.(2023)Gupta, Priyadarshi, and Gupta}]{gupta2023hateful}
Shrey Gupta, Pratyush Priyadarshi, and Manish Gupta. 2023.
\newblock Hateful comment detection and hate target type prediction for video comments.
\newblock In \emph{Proceedings of the 32nd ACM International Conference on Information and Knowledge Management}, pages 3923--3927.

\bibitem[{Kennedy et~al.(2022)Kennedy, Atari, Davani, Yeh, Omrani, Kim, Coombs~Jr, Havaldar, Portillo-Wightman, Gonzalez et~al.}]{kennedy2022introducing}
Brendan Kennedy, Mohammad Atari, Aida~Mostafazadeh Davani, Leigh Yeh, Ali Omrani, Yehsong Kim, Kris Coombs~Jr, Shreya Havaldar, Gwenyth Portillo-Wightman, Elaine Gonzalez, et~al. 2022.
\newblock Introducing the gab hate corpus: defining and applying hate-based rhetoric to social media posts at scale.
\newblock \emph{Language Resources and Evaluation}, 56(1):79--108.

\bibitem[{Kurki and Rask(2026)}]{kurki2026hyper}
Tuuli Kurki and Shadia Rask. 2026.
\newblock Hyper (in) visibility of blackness in spaces of whiteness: representations of race, gender and mental distress in the imageries of youth mental health services.
\newblock \emph{Journal of Youth Studies}, 29(3):421--442.

\bibitem[{Kurrek et~al.(2020)Kurrek, Saleem, and Ruths}]{kurrek2020towards}
Jana Kurrek, Haji~Mohammad Saleem, and Derek Ruths. 2020.
\newblock Towards a comprehensive taxonomy and large-scale annotated corpus for online slur usage.
\newblock In \emph{Proceedings of the Fourth Workshop on Online Abuse and Harms}, pages 138--149.

\bibitem[{Mathew et~al.(2021)Mathew, Saha, Yimam, Biemann, Goyal, and Mukherjee}]{mathew2021hatexplain}
Binny Mathew, Punyajoy Saha, Seid~Muhie Yimam, Chris Biemann, Pawan Goyal, and Animesh Mukherjee. 2021.
\newblock Hatexplain: A benchmark dataset for explainable hate speech detection.
\newblock In \emph{Proceedings of the AAAI conference on artificial intelligence}, volume~35, pages 14867--14875.

\bibitem[{Mody et~al.(2023)Mody, Huang, and De~Oliveira}]{mody2023curated}
Devansh Mody, YiDong Huang, and Thiago Eustaquio~Alves De~Oliveira. 2023.
\newblock A curated dataset for hate speech detection on social media text.
\newblock \emph{Data in Brief}, 46:108832.

\bibitem[{Mollas et~al.(2022)Mollas, Chrysopoulou, Karlos, and Tsoumakas}]{mollas2022ethos}
Ioannis Mollas, Zoe Chrysopoulou, Stamatis Karlos, and Grigorios Tsoumakas. 2022.
\newblock Ethos: a multi-label hate speech detection dataset.
\newblock \emph{Complex \& Intelligent Systems}, 8(6):4663--4678.

\bibitem[{Panchanadikar et~al.(2026)Panchanadikar, Hu, Guo, Hall, Hu, Vishwamitra, and Freeman}]{panchanadikar2026beyond}
Ruchi Panchanadikar, Yang Hu, Keyan Guo, Amelia~L Hall, Hongxin Hu, Nishant Vishwamitra, and Guo Freeman. 2026.
\newblock Beyond age-based restrictions: Rethinking children's online safety through comparing parent--child perspectives of risks in user-generated content games.
\newblock In \emph{Proceedings of the 2026 CHI Conference on Human Factors in Computing Systems}, pages 1--18.

\bibitem[{Pavlopoulos et~al.(2020)Pavlopoulos, Sorensen, Dixon, Thain, and Androutsopoulos}]{pavlopoulos2020toxicity}
John Pavlopoulos, Jeffrey Sorensen, Lucas Dixon, Nithum Thain, and Ion Androutsopoulos. 2020.
\newblock Toxicity detection: Does context really matter?
\newblock In \emph{Proceedings of the 58th annual meeting of the association for computational linguistics}, pages 4296--4305.

\bibitem[{Sachdeva et~al.(2022)Sachdeva, Barreto, Bacon, Sahn, Von~Vacano, and Kennedy}]{sachdeva2022measuring}
Pratik Sachdeva, Renata Barreto, Geoff Bacon, Alexander Sahn, Claudia Von~Vacano, and Chris Kennedy. 2022.
\newblock The measuring hate speech corpus: Leveraging rasch measurement theory for data perspectivism.
\newblock In \emph{Proceedings of the 1st Workshop on Perspectivist Approaches to NLP@ LREC2022}, pages 83--94.

\end{thebibliography}

\end{document}